\documentclass[twoside,11pt]{article}

\usepackage{blindtext}

\usepackage{dmlr2e}
\usepackage{amsmath}
\usepackage{graphicx}
\usepackage{multicol}
\usepackage{multirow}
\usepackage{arydshln}
\usepackage{tcolorbox}
\usepackage{booktabs}

\usepackage{lastpage}
\dmlrheading{23}{2025}{1-\pageref{LastPage}}{5/25; Revised 10/23}{11/23}{21-0000}{William Han, Chaojing Duan, Zhepeng Cen, Yihang Yao, Xiaoyu Song, Atharva Mhaskar, Dylan Leong, Michael A. Rosenberg, Emerson Liu, Ding Zhao} 
\def\openreview{\url{https://openreview.net/forum?id=XXXX}} 

\ShortHeadings{Signal, Image, or Symbolic}{Han, Duan, Cen, Yao, Song, Mhaskar, Leong, Rosenberg, Liu, and Zhao}
\firstpageno{1}

\begin{document}

\title{Signal, Image, or Symbolic: Exploring the Best Input Representation for Electrocardiogram-Language Models Through a Unified Framework}

\author{\name{William Han$^{1}$} \email{wjhan@andrew.cmu.edu}  \\
\name{Chaojing Duan$^{2}$} \email{chaojing.duan@ahn.org}  \\
\name{Zhepeng Cen$^{1}$} \email{zcen@andrew.cmu.edu}  \\
\name{Yihang Yao$^{1}$} \email{yihangya@andrew.cmu.edu}  \\
\name{Xiaoyu Song$^{1}$} \email{xiaoyuso@andrew.cmu.edu}  \\
\name{Atharva Mhaskar$^{1}$} \email{amhaskar@alumni.cmu.edu}  \\
\name{Dylan Leong$^{1}$} \email{dylanleo@andrew.cmu.edu}  \\
\name{Michael A. Rosenberg$^{3}$} \email{michael.a.rosenberg@cuanschutz.edu}  \\
\name{Emerson Liu$^{2}$} \email{emerson.liu@ahn.org}  \\
\name{Ding Zhao$^{1}$} \email{dingzhao@andrew.cmu.edu}  \\
\addr $^1$ Carnegie Mellon University \\
$^2$ Allegheny Health Network \\
$^3$ University of Colorado
}

\editor{My editor}

\maketitle

\begin{abstract}
Recent advances have increasingly applied large language models (LLMs) to electrocardiogram (ECG) interpretation, giving rise to Electrocardiogram–Language Models (ELMs). Conditioned on an ECG and a textual query, an ELM autoregressively generates a free-form textual response. Unlike traditional classification-based systems, ELMs emulate expert cardiac electrophysiologists by issuing diagnoses, analyzing waveform morphology, identifying contributing factors, and proposing patient-specific action plans. To realize this potential, researchers are curating instruction-tuning datasets that pair ECGs with textual dialogues and are training ELMs on these resources. Yet before scaling ELMs further, there is a fundamental question yet to be explored: What is the most effective ECG input representation? In recent works, three candidate representations have emerged—raw time-series signals, rendered images, and discretized symbolic sequences. We present the first comprehensive benchmark of these modalities across 6 public datasets and 5 evaluation metrics. We find symbolic representations achieve the greatest number of statistically significant wins over both signal and image inputs. We further ablate the LLM backbone, ECG duration, and token budget, and we evaluate robustness to signal perturbations. We hope that our findings offer clear guidance for selecting input representations when developing the next generation of ELMs. All code is available at \href{https://github.com/willxxy/ECG-Bench}{\texttt{github.com/willxxy/ECG-Bench}}.
\end{abstract}


\section{Introduction}
\label{intro}
Cardiovascular diseases (CVDs) claim approximately 18 million lives each year—a number that continues to rise \citep{worldhealthorganization_2024_cardiovascular}. Electrocardiograms (ECGs) are the primary tool for early CVD screening because they are accessible, noninvasive, and diagnostically valuable \citep{brief_history_cardio}. However, interpreting ECGs requires expert cardiac electrophysiologists to accurately analyze recordings, diagnose conditions, and recommend appropriate treatments. In parallel, the increasing demand for routine screenings and checkups \citep{10.1001/jamainternmed.2024.2270} exacerbates the national shortage of clinicians \citep{aamc_2024_the}, particularly in low-resource, rural areas \citep{johnson_2024_counties}.

To ease the burden on overworked specialists and improve screening efficiency, the deep learning community has increasingly focused on automating ECG analysis and diagnosis \citep{rajpurkar2017cardiologistlevelarrhythmiadetectionconvolutional, hannun_2019_cardiologistlevel, qiu-etal-2023-transfer, pmlr-v225-qiu23a}. Although most work has concentrated on classifying CVDs \citep{choi2023ecgbert, nonaka2020electrocardiogram_ecg_data_aug, Martin2021RealtimeFS, Strodthoff2021DeepLF}, recent research explores autoregressive generative approaches using large language models (LLMs). In this framework, given an ECG signal and a textual prompt (e.g., “Given my ECG, what conditions does it show?”), the model generates an appropriate textual response. We denote these as Electrocardiogram-Language Models (ELMs) \citep{zhao2024ecgchatlargeecglanguagemodel, lan2025gemempoweringmllmgrounded, liu2024teachmultimodalllmscomprehend, han2024ecgbytetokenizerendtoendgenerative}. ELMs offer enhanced flexibility and interpretability, mirroring the process by which cardiac electrophysiologists investigate ECGs.

Although ELMs are in their early stages and do not yet match the expertise of trained specialists, their potential is promising. With expanding public and private ECG datasets and recent insights into language model scaling laws \citep{kaplan2020scalinglawsneurallanguage}, developing expert-level ELMs is within reach. However, before significant computational resources are invested in training such models, a fundamental question must be addressed: \textbf{What is the optimal input representation for ECGs in ELMs?}

Recent works in Electrocardiogram-Language Models (ELMs) represent ECGs in three primary modalities: as raw signals \citep{zhao2024ecgchatlargeecglanguagemodel, tang2024electrocardiogramlanguagemodelfewshotquestion, tang2024electrocardiogramreportgenerationquestion}, images \citep{lan2025gemempoweringmllmgrounded, liu2024teachmultimodalllmscomprehend}, and symbols \citep{han2024ecgbytetokenizerendtoendgenerative}. In this work, we define each modality as follows:\\
\textbf{ECG Signal} The raw ECG signal is represented as a matrix $X_{\text{sig}} \in \mathbb{R}^{C \times L}$, where \( C \) denotes the number of leads (channels) and \( L \) is the number of time samples per lead. It is important to note in this work, all other modalities are derived from $X_{\text{sig}}$.\\
\textbf{ECG Image} An ECG image is derived from $X_{\text{sig}}$ via plotting and represented as a tensor $X_{\text{img}} \in \mathbb{R}^{H \times W \times C'}$, where \( H \) and \( W \) denote the image height and width, respectively, and \( C' \) is the number of color channels. We also create a synthetic three-channel version of $X_{\text{sig}}$, denoted \( X^*_{\text{sig}} \), by stacking \( X_{\text{sig}} \) three times along the channel dimension \( C' \) \citep{han2024ecgbytetokenizerendtoendgenerative}.\\
\textbf{ECG Symbol} We use ECG-Byte’s compression schema \citep{han2024ecgbytetokenizerendtoendgenerative} to convert ECG signals into symbols. First, a normalized and discretized ECG signal \(X_{\text{sig}}\) is mapped to a symbolic sequence using a set of symbols \(\mathcal{A} = \{\text{a}, \text{b}, \dots, \text{z}\}\). This sequence is then flattened into a one-dimensional array \(X_{\text{symb}} \in \mathcal{A}^{C\cdot L}\). Finally, a byte-pair encoding (BPE) process \citep{Gage1994ANA} compresses \(X_{\text{symb}}\) into a sequence of tokens from an extended vocabulary \(\mathcal{V}\), resulting in the final symbolic representation \(X_{\text{ID}} \in \mathcal{V}^{m}\), where \(m\) is the length of the token sequence.

Cardiac electrophysiologists typically analyze ECGs as images $X_{\text{img}}$, where signal values are plotted on paper. In contrast, deep learning applications for ECGs often process raw signal values directly because they are inherently informative \citep{na2024guidingmaskedrepresentationlearning, liu2024zeroshotecgclassificationmultimodal}. More recently, symbolic representations have gained attention in the more general time series and ECG domain \citep{tahery2024heartbertselfsupervisedecgembedding, han2024interpretationintracardiacelectrogramstextual, carson2024quantizedsymbolictimeseries, carson2024llmabbaunderstandtimeseries, elsworth2020abbaadaptivebrownianbridgebased}. Naturally, these symbolic representations can be formulated as discrete tokens, much like text. Although expert humans may interpret ECGs as images, \textbf{the most optimal input representation in the context of generative, autoregressive ELMs remains unexplored.}

This work bridges the gap by employing rigorous, empirical evaluations under controlled training settings to reveal statistically significant differences in performance across varying ECG input representations. We summarize our contributions as the following:\\
\textbf{1.} We provide an exhaustive performance benchmark for all three ECG modalities under controlled conditions, revealing statistically significant differences in performance.\\
\textbf{2.} We conduct extensive ablation studies that highlight the performance of each input representation when varying the LLM, ECG length, and sequence length.\\
\textbf{3.} We open source a comprehensive, unified framework for benchmarking ELMs across 4 input representations, 6 datasets, 4 ECG-specific encoders, 3 general, pretrained encoders, and 3 LLMs.\\
We hope that our comprehensive analysis provides future ELM researchers with clear guidelines on selecting the most appropriate representations when training ELMs.

\section{Background and Related Works}
\label{rel_works}
\subsection{Why Electrocardiogram-Language Models as Opposed to Classification Models?}
Most ECG diagnosis algorithms rely on simplistic single- or multi-label classification without providing any context behind their predictions \citep{rajpurkar2017cardiologistlevelarrhythmiadetectionconvolutional, hannun_2019_cardiologistlevel, nonaka2020electrocardiogram_ecg_data_aug, Martin2021RealtimeFS, Strodthoff2021DeepLF}. In contrast, autoregressive, generative Electrocardiogram-Language Models (ELMs) enable deep learning systems to diagnose cardiovascular diseases (CVD) from ECGs in a manner that more closely mimics the approach of cardiac electrophysiologists. Rather than simply outputting probability scores for predefined classification labels, ELMs leverage internet-scale pretrained knowledge that can be fine-tuned to generate free-form textual responses conditioned on the given ECG signal and query \citep{zhao2024ecgchatlargeecglanguagemodel}. ELMs can extend beyond diagnosis by identifying waveform characteristics, potential patient demographic factors contributing to CVD, and more \citep{oh2023ecgqacomprehensivequestionanswering, lan2025gemempoweringmllmgrounded}. This flexibility not only reduces the workload for cardiac electrophysiologists—who currently must review thousands of ECG records—but also empowers non-domain experts in the clinical setting to engage with advanced ECG analysis.

\subsection{What is an Electrocardiogram-Language Model?}
In its simplest form, an Electrocardiogram-Language Model (ELM) takes an ECG signal $X$ and a corresponding text query $Q$ to generate a text response $S$. Here, $X$ may represent a transformed version of the original signal (e.g., $X_{\text{img}}$, $X_{\text{ID}}$, $X^*_{\text{sig}}$) or the raw signal $X_{\text{sig}}$. Some works focus on a single-turn setting involving an ECG, a query, and a response \citep{han2024ecgbytetokenizerendtoendgenerative}, while others explore multi-turn conversations \citep{liu2024teachmultimodalllmscomprehend, zhao2024ecgchatlargeecglanguagemodel}. In multi-turn interactions, the first turn typically includes the ECG, the query, and the corresponding response, whereas subsequent turns contain only queries and responses derived from the initial ECG. In this work, we consider both settings.

\subsection{Why is the input representation important for Electrocardiogram-Language Models?}
To harness the benefits of scaling laws for large models \citep{kaplan2020scalinglawsneurallanguage}, it is crucial to establish a standard input representation for ECGs that optimizes the scaling of Electrocardiogram-Language Models (ELMs). There are four standard representations for an ECG signal: $X_{\text{sig}}$, $X_{\text{img}}$, and $X_{\text{ID}}$. $X_{\text{sig}}$ is the most natural form of ECG data and is most commonly used in ECG deep learning applications \citep{hannun_2019_cardiologistlevel, rajpurkar2017cardiologistlevelarrhythmiadetectionconvolutional} as well as in ECG self-supervised learning methods \citep{na2024guidingmaskedrepresentationlearning, liu2024zeroshotecgclassificationmultimodal}. However, concerns have grown regarding the accessibility of raw signals, as low-resource communities often use ECG recording machines that neither capture accurate signal values nor provide the capability to export data for collection \citep{Sangha2021.09.22.21263926}. Consequently, recent works have employed image representations of ECG signals \citep{liu2024teachmultimodalllmscomprehend, lan2025gemempoweringmllmgrounded}, denoted as $X_{\text{img}}$. Although $X_{\text{img}}$ is the format used by cardiac electrophysiologists to analyze ECG recordings, it poses several challenges for deep learning systems \citep{ni2025harnessingvisionmodelstime}. While this approach has gained popularity in the general time series domain \citep{SEMENOGLOU202339, li2023timeseriesimagesvision, zhuang2024itthinkitsorted}, some studies speculate that design choices in signal plotting can significantly affect a model's ability to interpret the visualization. Furthermore, both these ECG input representations require that recent ELMs utilize a pretrained or from-scratch trained ECG encoder to compress the high-dimensional signal into a latent vector, which is then concatenated with the textual query $Q$. Due to this overhead, recent works have explored converting ECG signals into textual representations via quantization and byte pair encoding \citep{han2024ecgbytetokenizerendtoendgenerative, tahery2024heartbertselfsupervisedecgembedding}, enabling direct fine-tuning of the LLM for generation. We note that \(X^{*}_{\text{sig}}\) introduced in Section~\ref{intro} is not a conventional way to represent ECGs; however, empirical evaluations reported in \cite{han2024ecgbytetokenizerendtoendgenerative} show that it performs on par with the standard signal representation \(X_{\text{sig}}\). Each modality offers its own advantages and disadvantages, therefore, it is critical to determine which representation yields statistically significant performance improvements in text generation for the future scaling of ELMs. Lastly, there has been a recent work that explored ELMs utilizing both $X_{\text{sig}}$ and $X_{\text{img}}$ \citep{lan2025gemempoweringmllmgrounded}. In our study, we only consider ELMs that utilize only one out of the four input representations. 

\subsection{How are Electrocardiogram-Language Models trained?}
\label{elm_train}
We identify three general settings in which Electrocardiogram-Language Models (ELMs) are trained: \textbf{2-Stage Scratch} \citep{zhao2024ecgchatlargeecglanguagemodel}, \textbf{End-to-End LLaVA} \citep{liu2024teachmultimodalllmscomprehend}, and \textbf{End-to-End} \citep{han2024ecgbytetokenizerendtoendgenerative}. Below, we define each setting at a high level.

\paragraph{2-Stage Scratch}
Some works develop an ECG-specific encoder \( f_{\text{ECG}} : \mathbb{R}^{C \times L} \to \mathbb{R}^d \) using self-supervised learning (SSL) on the raw ECG signal \( X_{\text{sig}} \) \citep{na2024guidingmaskedrepresentationlearning, pham2024cmeltcontrastiveenhancedmasked, kiyasseh2021clocscontrastivelearningcardiac, choi2023ecgbert, vaid2022heartbeit, jin2025readingheartlearningecg}. Here, \( C \) represents the number of channels, \( L \) the number of time steps, and \( d \) the dimension of the latent space where ECG data is encoded. SSL approaches, such as masked image modeling or contrastive learning, are employed \citep{na2024guidingmaskedrepresentationlearning, liu2024zeroshotecgclassificationmultimodal, oh2022leadagnosticselfsupervisedlearninglocal}. In contrastive learning, a text encoder \( f_{\text{text}} : \mathbb{R} \to \mathbb{R}^d \) may map textual reports to the same \( d \)-dimensional latent space for alignment with ECG encodings. For simplicity, we refer to this initial training of the encoder as training \( f_{\text{ECG}} \), and denote this setting as \textbf{2-stage scratch}.

\paragraph{End-to-End LLaVA}
Other works utilize general pretrained image encoders, such as CLIP \citep{Radford2021LearningTV} or ViT \citep{dosovitskiy2021image}, as \( f_{\text{ECG}} \) \citep{han2024ecgbytetokenizerendtoendgenerative, pmlr-v225-qiu23a, liu2024teachmultimodalllmscomprehend, lan2025gemempoweringmllmgrounded}. Since these encoders expect image inputs, the ECG data must be adapted: either by creating a synthetic three-channel ECG signal \( X^*_{\text{sig}} \), or by using an image representation \( X_{\text{img}} \) of the ECG \citep{lan2025gemempoweringmllmgrounded, liu2024teachmultimodalllmscomprehend}. In the LLaVA-style approach \citep{liu2023visualinstructiontuning, liu2024teachmultimodalllmscomprehend}, \( f_{\text{ECG}} \) is frozen, and a learnable projection matrix \( W \in \mathbb{R}^{h \times d} \) is introduced, where \( h \) is the hidden dimension of the LLM. The latent vector \( z = f_{\text{ECG}}(X) \) is projected to \( z' = W z \), concatenated with the embedded query \( Q \), and fed into the LLM to generate the response \( S \). Only \( W \) and the LLM are trained, while \( f_{\text{ECG}} \) remains fixed. This setting is denoted as \textbf{End-to-End LLaVA}.

\paragraph{End-to-End} 
In the \textbf{End-to-End} training setting, the raw ECG signal \(X_{\text{sig}}\) is discretized into a sequence of token IDs, \(X_{\text{ID}}\), using ECG-Byte \citep{han2024ecgbytetokenizerendtoendgenerative}. ECG-Byte compresses the signal by first quantizing the amplitudes and then applying Byte Pair Encoding (BPE) \citep{Gage1994ANA}.
Therefore, one can directly train the LLM for autoregressive generation since $X_{\text{ID}}$, query $Q$, and response $S$ are all tokens.

\begin{figure}
    \centering
    \includegraphics[width=1\linewidth]{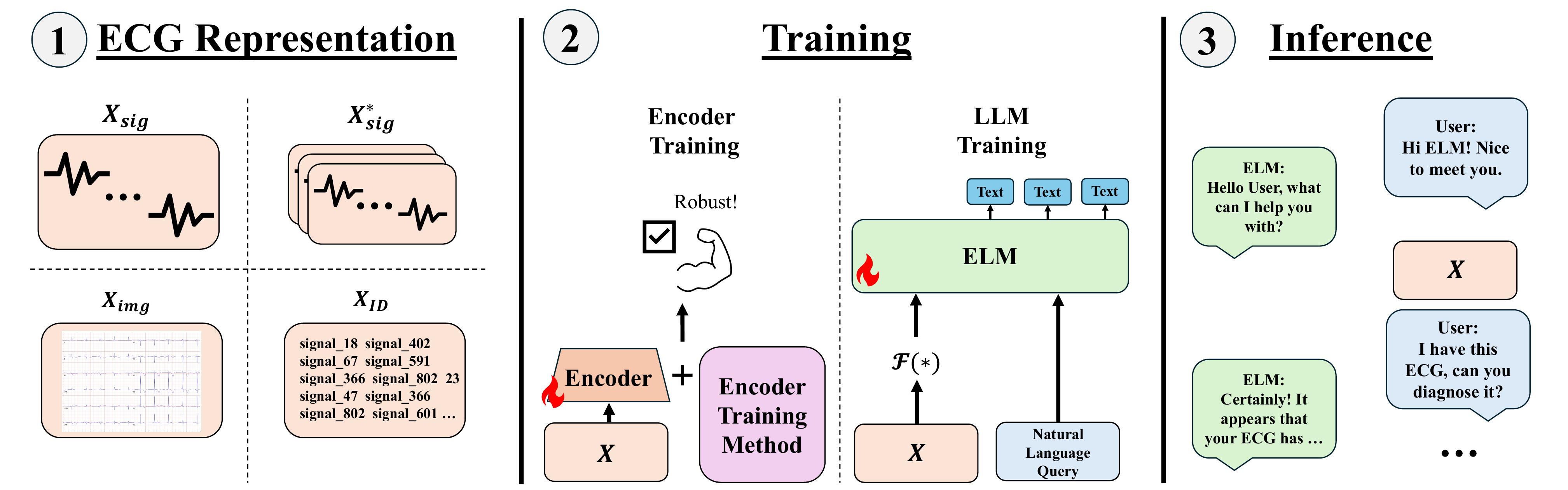}
    \caption{A high-level overview of our training and evaluation pipeline. The input data is represented as $X = \{X_{\text{sig}}, X^*_{\text{sig}}, X_{\text{img}}, X_{\text{ID}}\}$ as seen in Step 1. Step 2 comprises two modes: Encoder training and LLM training. During Encoder training, an ECG-specific encoder is trained from scratch. In LLM training, any method used to compress the ECG into an appropriate representation for the ELM is denoted by $\mathcal{F}(*)$ for simplicity; this mode covers all training methods described in Subsection~\ref{elm_train}. In Step 3, inference is performed using a conversational template applicable to both single- and multi-turn settings.}
    \label{fig:ts_img_sig}
\end{figure}

\section{Methods}
\subsection{Datasets}
We categorize the datasets used into the following: Base and Mapping datasets. Base datasets refer to the original ECG signal datasets in which Mapping datasets are derived from. Mapping datasets map back to the signals provided in the Base datasets, however, they contain additional features, most notably questions, answers, and conversations. For Base datasets, we consider MIMIC-IV-ECG \citep{johnson_2023_mimiciv}, PTB-XL \citep{wagner_ptb-xl_2020}, Code-15\% \citep{ribeiro_2021_code15}, CSN \citep{zheng_2020_a}, and CPSC \citep{andres_2022_classification}. For Mapping datasets, we consider the PULSE ECG-Instruct \citep{liu2024teachmultimodalllmscomprehend}, PULSE ECG-Bench \citep{liu2024teachmultimodalllmscomprehend}, ECG-Chat Pretrain \citep{zhao2024ecgchatlargeecglanguagemodel}, ECG-Chat Instruct \citep{zhao2024ecgchatlargeecglanguagemodel}, ECG-QA \citep{oh2023ecgqacomprehensivequestionanswering} (both PTB-XL and MIMIC-IV-ECG versions) datasets. For ECG-QA, we only consider the single-verify, single-choose, and single-query categorized questions. ECG-Chat Pretrain and ECG-QA datasets exclusively contain only single-turn question and answers, while all other Mapping datasets consider both single- and multi-turn conversations.

For \textbf{2-Stage Scratch}, we train \( f_{\text{ECG}} \) using ECG signals (and corresponding textual clinical diagnostic reports if needed for some contrastive learning approaches) from the Base MIMIC-IV-ECG dataset. For \textbf{End-to-End LLaVA} and \textbf{End-to-End} we directly train on the Mapping datasets.

\subsection{ECG Signal Preprocessing}
\label{ecg_preprocess}
We preprocess all ECG signals from the Base datasets considered in this study the same way for comparability, largely following \citep{han2024ecgbytetokenizerendtoendgenerative}. We first adjust all lead arrangements to be consistent with PTB-XL (i.e., [I, II, III, aVL, aVR, aVF, V1–V6]). Powerline noise is removed using bidirectional notch filters at 50 Hz and 60 Hz with a quality factor (Q) of 30. A fourth-order Butterworth bandpass filter (0.5–100 Hz) is applied to isolate the ECG signal, while a bidirectional highpass filter (cutoff 0.05 Hz) is used to reduce baseline drift. Daubechies-6 (db6) wavelet denoising at level 4 is then applied, utilizing a soft threshold based on the median absolute deviation of the detail coefficients. Due to varying signal frequencies across the Base datasets, we unify all frequencies to 250 Hz. We utilize non-overlapping 5-second segments for all settings except when training the ECG-Byte tokenizer for \textbf{End-to-End}, as the authors suggest utilizing the full, unsegmented ECG recording \citep{han2024ecgbytetokenizerendtoendgenerative}. 

\subsection{ECG Signal Transformations}
We provide the detailed method in which we transform and prepare each input representation from the original ECG signal $X_{\text{sig}}$.

\paragraph{ECG Signal}
As mentioned before, we define the raw ECG signal as a 2-dimensional matrix $X_{\text{sig}} \in \mathbb{R}^{C \times L}$, where \( C \) denotes the number of leads (channels) and \( L \) is the number of time samples per lead. We also consider the stacked ECG signal representation $X^*_{\text{sig}}$. We preprocess both $X_{\text{sig}}$ and $X^*_{\text{sig}}$ as described in Subsection~\ref{ecg_preprocess} and utilize the preprocessed $X_{\text{sig}}$ or $X^*_{\text{sig}}$ as input to the ELM. We normalize to [0, 1] by using the minimum and maximum values of the given ECG signal.

\paragraph{ECG Image}
From the preprocessed $X_{\text{sig}}$ described above, we generate a plotted visualization of all 12 leads, denoted as $X_{\text{img}}$, to mimic the display observed by human cardiac electrophysiologists (see Figure~\ref{fig:ts_img_sig}). Previous works have employed the ECG-Image-Kit \citep{shivashankara2024ecgimagekitsyntheticimagegeneration} to create realistic, synthetic $X_{\text{img}}$ using various in-house augmentations that simulate wrinkles, handwriting, and other effects \citep{liu2024teachmultimodalllmscomprehend, lan2025gemempoweringmllmgrounded}. In contrast, our study reports all results without any augmentations unless otherwise specified, as we aim to maintain a controlled setting by utilizing only the primary transformation from $X_{\text{sig}}$ to $X_{\text{img}}$. Additionally, to simulate recordings directly obtained from an ECG recording machine, we do not normalize the ECG prior to plotting.

\paragraph{ECG Symbol}
From the preprocessed $X_{\text{sig}}$, we utilize ECG-Byte \citep{han2024ecgbytetokenizerendtoendgenerative} to create a symbolic representation by quantizing continuous ECG signal values into discrete tokens, denoted as $X_{\text{ID}}$. In our study, we train ECG-Byte using 5000 merges on 300,000 randomly sampled, unsegmented, preprocessed ECG signals. Notably, our approach deviates from the original ECG-Byte methodology \cite{han2024ecgbytetokenizerendtoendgenerative}. Instead of normalizing the ECGs using global percentiles, we employ instance normalization, where each ECG is scaled individually by computing its minimum and maximum values to map the data to the range [0, 1].

\subsection{ECG Encoders}
We consider a wide range of ECG encoders $f_{\text{ECG}}$ to capture a variety of state of the art self-supervision methods for ECG signals $X_{\text{sig}}$ and general, pretrained encoders for $X_{\text{img}}$ and $X^*_{\text{sig}}$ as well. For self-supervision methods, we utilize popular ECG encoders namely MERL \citep{liu2024zeroshotecgclassificationmultimodal}, ST-MEM \citep{na2024guidingmaskedrepresentationlearning}, MLAE \citep{9980411}, and MTAE \citep{9980411}. For general, pretrained encoders, we utilize CLIP \citep{Radford2021LearningTV}, ViT \citep{dosovitskiy2021image}, and SigLIP \citep{zhai2023sigmoidlosslanguageimage}.

\subsection{Large Language Models}
In this study, we utilize the Llama 3.2 1B Instruct \citep{grattafiori2024llama3herdmodels} checkpoint from the HuggingFace API \citep{wolf2020huggingfaces} unless specified otherwise. We also fix the token sequence length $T$ to 1024 unless specified otherwise. We present ablation studies in Subsection~\ref{ablation} of using other similar size models such as Gemma 2 2B Instruct \citep{gemmateam2024gemma2improvingopen} and Qwen 2.5 1.5B Instruct \citep{qwen2025qwen25technicalreport} for each input representation and training paradigm as well as different sequence lengths $T$. We note that Gemma 2B Instruct does not have a system prompt $q_{\text{sys}}$ integrated into their chat template, therefore we do not include $q_{\text{sys}}$.

\subsection{Padding Schema}
\label{sec:pad}
We outline the distinct padding strategies employed for each training paradigm to construct the input sequence
\(Y = (y_1, y_2, \dots, y_T)\), where \(T\) denotes the maximum sequence length.

For \textbf{2-Stage Scratch} and \textbf{End-to-End LLaVA}, the signal is represented by a single token \(\langle\text{signal}\rangle\), while the remaining components of \(Y\) are text tokens (system prompt, user query, assistant response).  
Since the signal occupies only one position, we adjust the text tokens to fit the sequence length \(T\). If the total length of the text tokens plus the signal token is \(\,<T\), we pad on the left with pad tokens until the sequence reaches \(T\). If the total length exceeds \(T\), we truncate the sequence to the first \(T\) tokens.
Thus, padding or truncation is applied solely to the text part, and the final sequence length is always \(T\).

In the \textbf{End-to-End} training method, the signal is a sequence of tokens \(X_{\text{ID}}\), concatenated with text tokens to form \(Y\).
To manage the variable-length \(X_{\text{ID}}\) within \(T\), we truncate \(X_{\text{ID}}\) so that it fits within \(T\) together with the text tokens, but always keep at least \(\ge 500\) signal tokens. If the combined length of the truncated signal and text tokens still exceeds \(T\) we truncate the text tokens so the total length equals \(T\). If the total length is \(<T\), we pad on the left with pad tokens to reach \(T\).
This scheme prioritizes retaining a substantial portion of the signal while still fitting the text context into the fixed sequence length \(T\).

\subsection{Learning Objective}
We consider an autoregressive (next-token prediction) training objective that is applicable to any conversation input format. In our general formulation, the complete input sequence is given by $Y = \bigl( y_1, y_2, \dots, y_T \bigr)$,
which is constructed by concatenating various components such as a system prompt \( q_{\text{sys}} \), a signal placeholder \( <\text{signal}> \) for the projected latent vector \( z' \) in \textbf{2-Stage Scratch} and \textbf{End-to-End LLaVA} or a sequence of signal tokens \( X_{\text{ID}} \) for \textbf{End-to-End}, the first user query \( q_1 \) (always coupled with the signal), and subsequent query-response pairs $q_1, s_1, q_2, s_2, \dots, q_n, s_n$.
A representative example of the token template is given in Appendix~\ref{input_temp}.

To ensure that training focuses only on the desired targets—specifically, the assistant responses and their corresponding end-of-turn tokens—we define a target labeling function
\[
\ell: \{1, 2, \dots, T\} \to \mathcal{V} \cup \{-100\},
\]
where \( \mathcal{V} \) is the extended vocabulary set, as follows:
\[
\ell(t) = \begin{cases}
y_t, & \text{if } y_t \text{ is part of } s_i  \text{ or is an end-of-turn token (e.g., } <|\text{eot\_id}|>\text{)}, \\
-100, & \text{otherwise}.
\end{cases}
\]
Here, the value \( -100 \) is used to mask out tokens that are not part of the target output during loss computation.

Then, if we define the target positions as
\[
\mathcal{T} = \{ t \in \{1, \dots, T\} : \ell(t) \neq -100 \},
\]
the loss becomes
\[
\mathcal{L}(\theta) = - \sum_{t \in \mathcal{T}} \log p_\theta\bigl( y_t \mid y_{<t} \bigr).
\]

This formulation is agnostic to specific conversation templates and ensures that only the tokens corresponding to the assistant responses \( S = \{ s_1, s_2, \dots, s_n \} \) and end-of-turn tokens contribute to the loss, following the approach in LLaVA \citep{liu2023visualinstructiontuning}.

\begin{figure}
    \centering
    \includegraphics[width=1\linewidth]{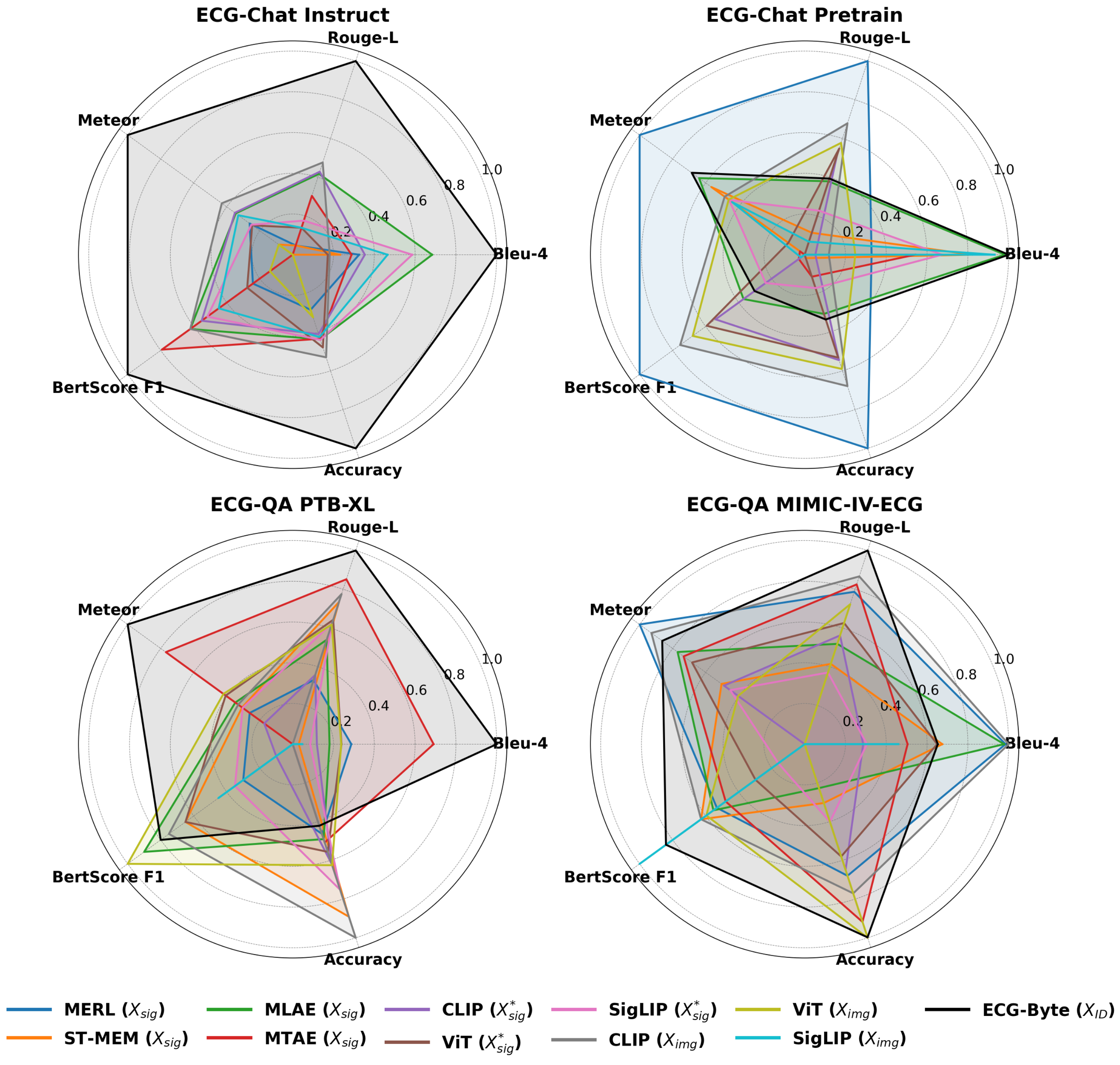}
    \caption{Spider charts for the performance of each model and training paradigm. We want to note that $X_{\text{sig}}$, $X^*_{\text{sig}}$, $X_{\text{img}}$, and $X_{\text{ID}}$ utilize \textbf{2-Stage Scratch}, \textbf{End-to-End LLaVA}, \textbf{End-to-End LLaVA}, and \textbf{End-to-End} training paradigms respectively. We include a table representing the same results in Table~\ref{tab:main} of Appendix~\ref{apd:results}.}
    \label{fig:main_results}
\end{figure}

\section{Experiments}
\subsection{Experimental Settings}
All training and inference settings are uniform across input representations and methods of training ELMs. We use the AdamW optimizer \citep{kingma2017adam} with a learning rate of 1e-4 and a weight decay of 1e-2 \citep{NIPS1991_8eefcfdf}, and we adopt the custom learning rate scheduler described in ECG-Byte \citep{han2024ecgbytetokenizerendtoendgenerative}. For the first stage of \textbf{2-Stage Scratch}, we train on a subset of 800,000 ECGs drawn from the MIMIC-IV-ECG dataset \citep{johnson_2023_mimiciv}. For training the LLM, we first partition the data into training and test sets in a 70:10 ratio. Due to computational resource limitations, we then sample 400,000 instances from the training set and 20,000 instances from the test set. In cases where some Mapping datasets contain fewer instances than these sampled counts, we use the full train and test sets from the initial split. When fine-tuning the LLM, we employ parameter-efficient fine-tuning with LoRA \citep{hu2021loralowrankadaptationlarge}, setting $\text{rank} = 16$, $\alpha_{\text{LoRA}} = 32$, and $\text{dropout} = 0.05$. We also apply Flash Attention 2 \citep{dao2023flashattention2fasterattentionbetter} when training the LLM. During inference, we employ widely used text generation metrics, including BLEU-4 \citep{Papineni2002BleuAM}, ROUGE-L \citep{Lin2004ROUGEAP}, METEOR \citep{Banerjee2005METEORAA}, and BertScore F1 \citep{Zhang2020BERTScoreET}. We additionally report generation accuracy, defined as the proportion of cases in which the generated response exactly matches the ground truth. All reported results are presented as averages with standard deviations computed over 5 random seeds. All experiments are conducted on 2 A5000 (24 GB) and 2 A6000 (48 GB) NVIDIA GPUs.

\section{Results}
In the following subsections, we analyze the performance of each input representation across multiple training paradigms. We begin with the main results on four datasets, highlighting statistically significant differences, then report performance on the PULSE ECG‑Bench benchmark. We also assess the robustness of each representation under signal perturbations. We then go into a series of ablation studies to highlight performance differences under model component variations.

\subsection{Main Results}
We summarize our main results across four datasets in Figure~\ref{fig:main_results} using a radar (spider) chart, where each axis corresponds to a normalized evaluation metric: BLEU-4, ROUGE-L, METEOR, BertScore F1, and Accuracy. Metrics are normalized to a common scale of [0,1] to enable consistent comparison across heterogeneous scales. For each dataset, results are averaged across five random seeds, with standard deviations reported in Table~\ref{tab:main} in Appendix~\ref{apd:results}. Each model configuration is represented using a distinct color and label. We denote input modality variants as $X_{\text{sig}}$, $X^*_{\text{sig}}$, $X_{\text{img}}$, and $X_{\text{id}}$, corresponding to \textbf{2-Stage Scratch}, \textbf{End-to-End LLaVA}, \textbf{End-to-End LLaVA}, and \textbf{End-to-End} training, respectively.

For each (dataset, metric) pair we identify the top‐scoring model and its runner-up and perform a paired \(t\)-test across the same five seeds.  Let the per-seed difference be \(d_i\).  Then
\[
\bar d = \frac{1}{n}\sum_{i=1}^{n} d_i,\quad
s_d = \sqrt{\frac{1}{n-1}\sum_{i=1}^{n} (d_i - \bar d)^2},\quad
t = \frac{\bar d}{s_d / \sqrt{n}},\quad n = 5,
\]
\[
p = 2\!\left[1 - F_{t}(|t|; n-1)\right],
\]
where \(F_{t}(\cdot; \nu)\) is the CDF of Student’s \(t\)-distribution with \(\nu = n-1\) degrees of freedom.  Using a two-tailed \(\alpha = 0.05\), we count a “significant win” when \(p < 0.05\).  Aggregating wins across all tests (Table~\ref{tab:overall_wins}), the \textbf{End-to-End ECG-Byte} variant (\(X_{\text{id}}\)) dominates, with the \textbf{2-Stage Scratch} MERL model placing second.

\begin{table}[!htp]\centering
\caption{Count of statistically significant metric wins across all datasets.}
\label{tab:overall_wins}
\begin{tabular}{l c}
\toprule
Model & \#\,Significant Wins\\
\midrule
ECG‑Byte   & \textbf{9} \\
MERL       & 4 \\
CLIP ($X_\text{img}$)  & 1 \\
SigLIP ($X_\text{img}$)& 1 \\
\bottomrule
\end{tabular}
\end{table}

\begin{table}[!htp]\centering
\caption{Mean results with standard deviations over five random seeds when training on the PULSE ECG-Instruct dataset and zero-shot inferencing on the PULSE ECG-Bench dataset.}\label{tab:ecg-bench}
{\resizebox{\columnwidth}{!}{%
\begin{tabular}{ccccccccc}\toprule
Representation & Training Paradigm & $\mathcal{F}(*)$ & Bleu-4 & Rouge-L & Meteor & BertScore F1 & Accuracy \\\midrule
\multirow{4}{*}{$X_{\text{sig}}$} & \multirow{4}{*}{2-Stage Scratch} & MERL   & 1.25 ± 0.11 & 18.51 ± 0.53 & 11.67 ± 0.29 & 93.75 ± 0.03 & 14.66 ± 0.54 \\
                                    &                               & ST-MEM & 1.19 ± 0.10 & 15.43 ± 0.49 &  9.98 ± 0.24 & 93.45 ± 0.04 & 11.05 ± 0.51 \\
                                    &                               & MLAE   & 1.16 ± 0.12 & 18.80 ± 0.34 & 11.21 ± 0.22 & 94.58 ± 0.03 & 16.08 ± 0.33 \\
                                    &                               & MTAE   & 0.81 ± 0.04 & 12.42 ± 0.23 &  8.28 ± 0.15 & 92.54 ± 0.03 &  7.97 ± 0.32 \\
\midrule
\multirow{3}{*}{$X^*_{\text{sig}}$} & \multirow{3}{*}{End-to-End LLaVA} & CLIP   & 0.88 ± 0.04 & 15.23 ± 0.35 &  9.18 ± 0.19 & 93.42 ± 0.05 & 12.90 ± 0.35 \\
                                    &                               & ViT    & 0.95 ± 0.06 & 17.88 ± 0.60 & 10.40 ± 0.36 & 94.57 ± 0.01 & 15.57 ± 0.57 \\
                                    &                               & SigLIP & 1.29 ± 0.03 & 18.51 ± 0.51 & 11.40 ± 0.23 & 94.25 ± 0.04 & 15.02 ± 0.53 \\
\midrule
\multirow{3}{*}{$X_{\text{img}}$}   & \multirow{3}{*}{End-to-End LLaVA} & CLIP   & 1.44 ± 0.16 & 18.96 ± 0.40 & 11.52 ± 0.25 & 93.97 ± 0.04 & 16.20 ± 0.33 \\
                                    &                               & ViT    & 0.93 ± 0.08 & 18.35 ± 0.61 & 10.43 ± 0.31 & 94.55 ± 0.03 & 16.57 ± 0.63 \\
                                    &                               & SigLIP & 0.93 ± 0.11 & 17.70 ± 0.24 & 10.26 ± 0.14 & 93.91 ± 0.03 & 15.46 ± 0.27 \\
\midrule
$X_{\text{id}}$                     & End-to-End                     & ECG-Byte & \textbf{1.49 ± 0.20} & \textbf{20.35 ± 0.62} & \textbf{12.21 ± 0.44} & \textbf{94.60 ± 0.04} & \textbf{17.08 ± 0.48} \\
\bottomrule
\end{tabular}}}
\end{table}

\subsection{Results on PULSE ECG-Bench}
Table~\ref{tab:ecg-bench} presents the zero-shot performance of models trained on the PULSE ECG-Instruct dataset when evaluated on the PULSE ECG-Bench benchmark. Across all metrics and representation modalities, ECG-Byte ($X_{\text{id}}$), trained \textbf{End-to-End}, achieves the highest scores, demonstrating generalization to unseen samples. 

\begin{table*}[!ht]
\centering
\caption{Performance change ($\Delta$ = Perturbed – Baseline) averaged over 5 random seeds on the ECG-Chat Instruct dataset.}
\label{tab:ecg_chat_instruct_Perturbed}
\resizebox{\textwidth}{!}{%
\begin{tabular}{lll*{5}{c}}
\toprule
\multirow{2}{*}{Representation}
  & \multirow{2}{*}{Training Paradigm}
  & \multirow{2}{*}{$\mathcal{F}(*)$}
  & \multicolumn{5}{c}{$\Delta$ (Perturbed – Baseline)} \\
\cmidrule(lr){4-8}
 & & & BLEU-4 & ROUGE-L & METEOR & BERTScore F1 & Accuracy \\
\midrule
\multirow{4}{*}{$X_{\text{sig}}$}
 & \multirow{4}{*}{2-Stage Scratch}
 & MERL  & -0.07 & -0.05 & -0.08 & -0.08 & +0.05 \\[2pt]
 & & ST-MEM & +0.04 & -0.03 & -0.01 & -0.01 & +0.01 \\[2pt]
 & & MLAE  & +0.01 & -0.07 & -0.05 & -0.09 & -0.09 \\[2pt]
 & & MTAE  & -0.04 & \textbf{+0.03} & \textbf{+0.04} & -0.04 & +0.10 \\
\cmidrule(r){1-8}
\multirow{3}{*}{$X^{*}_{\text{sig}}$}
 & \multirow{3}{*}{End-to-End LLaVA}
 & CLIP  & +0.03 & 0.00 & -0.03 & \textbf{0.00} & +0.11 \\[2pt]
 & & ViT   & +0.05 & 0.00 & +0.03 & \textbf{0.00} & +0.08 \\[2pt]
 & & SigLIP & +0.07 & -0.04 & -0.08 & -0.01 & \textbf{+0.13} \\
\cmidrule(r){1-8}
\multirow{3}{*}{$X_{\text{img}}$}
 & \multirow{3}{*}{End-to-End LLaVA}
 & CLIP  & -0.10 & -0.02 & -0.06 & \textbf{0.00} & -0.12 \\[2pt]
 & & ViT   & -0.01 & -0.03 & -0.08 & -0.01 & -0.06 \\[2pt]
 & & SigLIP & -0.01 & -0.03 & +0.02 & \textbf{0.00} & -0.04 \\
\cmidrule(r){1-8}
$X_{\text{id}}$ & End-to-End & ECG-Byte
 & \textbf{+0.09} & -0.03 & +0.02 & \textbf{0.00} & -0.05 \\
\bottomrule
\end{tabular}}
\end{table*}

\subsection{Robustness to ECG Perturbations}
To probe inference‑time stability, we randomly perturb each ECG signal. Concretely, every example has a $50\%$ chance of being altered; if selected, we add zero‑mean Gaussian noise with standard deviation $0.05\,\sigma_{x}$, where $\sigma_{x}$ is the signal’s sample standard deviation. With a further independent $50\%$ probability we then superimpose a low‑frequency sinusoidal baseline‑wander of amplitude $0.07\,\|x\|_{\infty}$ and random frequency chosen uniformly from $[1,5]\pi$ radians across the record length. This two‑stage process yields a lightly perturbed yet physiologically plausible waveform.

Table~\ref{tab:ecg_chat_instruct_Perturbed} reports deltas on original versus perturbed inputs. We report the numerical performance metrics in Table~\ref{tab:ecg_chat_instruct_Perturbed_main} of Appendix~\ref{apd:results}. Across all representations the average degradation is minor, suggesting that the language model component derives limited utility from the raw signal—an observation consistent with the attention visualizations reported by \citet{han2024ecgbytetokenizerendtoendgenerative}, where weights concentrate on the textual portion of the prompt. These results underline the need for future architectures and training strategies that compel the model to leverage ECG‑level information more effectively.

\subsection{Ablation Study}
\label{ablation}
We conduct ablation studies on utilizing different LLMs, sequence lengths $T$, ECG lengths $L$ to expose more insights on how ablating these parameters affects the overall performance. With the exception of the ablating parameter, we utilize the Llama 3.2 1B Instruct LLM, $T=1024$, and $L=1250$. Lastly, we report results on the same train and test split from the Mapping dataset ECG-Chat Instruct utilized in Figure~\ref{fig:main_results} since this dataset comprises of both single- and multi-turn conversations.

\newpage
\begin{table}[!htp]\centering
\caption{Mean results over 5 random seeds with standard deviations when varying LLMs.}\label{tab:llm}
{\resizebox{\columnwidth}{!}{%
\begin{tabular}{lccccccccc}\toprule
LLMs &Representation &Training Paradigm &$\mathcal{F}(*)$ &Bleu-4 &Rouge-L &Meteor &BertScore F1 &Accuracy \\\midrule
\multirow{11}{*}{Gemma 2B Instruct} &\multirow{4}{*}{$X_{\text{sig}}$} &\multirow{4}{*}{2-Stage Scratch} &MERL & 17.26 ± 0.03& 61.42 ± 0.04&54.54 ± 0.02 & 94.93 ± 0.01 & 7.89 ± 0.11\\
& & &ST-MEM& 16.44 ± 0.04&61.02 ± 0.04 &54.03 ± 0.08 &94.92 ± 0.01 &6.98 ± 0.13 \\
& & &MLAE & 16.57 ± 0.01& 62.45 ± 0.06&54.13 ± 0.07 & 95.08 ± 0.00& 7.63 ± 0.11\\
& & &MTAE & 15.49 ± 0.03&61.18 ± 0.01 & 54.68 ± 0.02& 94.91 ± 0.01& 6.60 ± 0.08\\
\cmidrule(r){2-9} 
&\multirow{3}{*}{$X^*_{\text{sig}}$} &\multirow{3}{*}{End-to-End LLaVA} &CLIP& 15.55 ± 0.04& 59.88 ± 0.06& 51.80 ± 0.05& 94.52 ± 0.02& 7.12 ± 0.09\\
& & &ViT& 15.54 ± 0.04& 60.01 ± 0.04& 52.01 ± 0.06& 95.01 ± 0.00&7.24 ± 0.07 \\
& & & SigLIP& 16.01 ± 0.02& 61.28 ± 0.05&53.84 ± 0.07 &95.21 ± 0.00 &7.28 ± 0.12 \\
\cmidrule(r){2-9} 
&\multirow{3}{*}{$X_{\text{img}}$} &\multirow{3}{*}{End-to-End LLaVA} &CLIP& 16.72 ± 0.06& 61.69 ± 0.07& 52.98 ± 0.10& 94.36 ± 0.03& 7.29 ± 0.03\\
& & &ViT& 15.72 ± 0.01&59.66 ± 0.04 & 51.91 ± 0.02& 94.73 ± 0.01& 6.08 ± 0.12\\
& & & SigLIP& 17.16 ± 0.04& 61.92 ± 0.05& 53.25 ± 0.08& 95.58 ± 0.01& 7.52 ± 0.09 \\
\cmidrule(r){2-9} 
&$X_{\text{ID}}$ &End-to-End &ECG-Byte & \textbf{20.59 ± 0.04}& \textbf{65.39 ± 0.03}& \textbf{57.83 ± 0.09}& \textbf{96.12 ± 0.01}& \textbf{10.56 ± 0.06}\\
\midrule
\multirow{11}{*}{Qwen 2.5 1.5B Instruct} &\multirow{4}{*}{$X_{\text{sig}}$} &\multirow{4}{*}{2-Stage Scratch} &MERL & 16.12 ± 0.04&60.33 ± 0.05 &53.87 ± 0.07 &95.21 ± 0.01 &6.89 ± 0.06 \\
& & &ST-MEM& 15.42 ± 0.02& 58.63 ± 0.06& 50.74 ± 0.05&95.74 ± 0.02 &6.52 ± 0.03 \\
& & &MLAE & 15.34 ± 0.03& 58.51 ± 0.06&50.86 ± 0.05 &95.23 ± 0.01 &6.53 ± 0.05 \\
& & &MTAE & 15.49 ± 0.08& 59.48 ± 0.03&51.36 ± 0.04 &95.28 ± 0.01 &6.72 ± 0.07 \\
\cmidrule(r){2-9} 
&\multirow{3}{*}{$X^*_{\text{sig}}$} &\multirow{3}{*}{End-to-End LLaVA} &CLIP& 14.97 ± 0.01&59.89 ± 0.03 &53.08 ± 0.02 & 95.07 ± 0.01& 5.02 ± 0.10\\
& & &ViT& 13.65 ± 0.04& 58.81 ± 0.03 &52.14 ± 0.05 & 95.01 ± 0.00&5.01 ± 0.08 \\
& & & SigLIP& 16.51 ± 0.06& 61.47 ± 0.07&53.92 ± 0.07 &95.19 ± 0.01 &6.87 ± 0.03 \\
\cmidrule(r){2-9} 
&\multirow{3}{*}{$X_{\text{img}}$} &\multirow{3}{*}{End-to-End LLaVA} &CLIP& 15.88 ± 0.01& 60.33 ± 0.01& 53.08 ± 0.01&95.14 ± 0.01 &5.45 ± 0.03 \\
& & &ViT& 15.03 ± 0.03&60.35 ± 0.02 &52.99 ± 0.06 & 95.08 ± 0.00&5.04 ± 0.04 \\
& & & SigLIP& 16.28 ± 0.01& 61.80 ± 0.05&54.41 ± 0.03 & 95.25 ± 0.01& 7.03 ± 0.00\\
\cmidrule(r){2-9} 
&$X_{\text{ID}}$ &End-to-End &ECG-Byte & \textbf{19.02 ± 0.05}& \textbf{65.61 ± 0.02}&\textbf{58.31 ±0.08} & \textbf{97.87 ± 0.01}& \textbf{9.45 ± 0.05}\\
\midrule
\multirow{11}{*}{Llama 3.2 1B Instruct} &\multirow{4}{*}{$X_{\text{sig}}$} &\multirow{4}{*}{2-Stage Scratch} &MERL &22.29 ± 0.04 &65.26 ± 0.05 & 58.52 ± 0.05& 96.08 ± 0.01&8.70 ± 0.07 \\
& & &ST-MEM& 22.09 ± 0.03& 65.26 ± 0.02 &57.83 ± 0.02 &96.01 ± 0.00 &7.75 ± 0.05 \\
& & &MLAE & 23.06 ± 0.04 &66.27 ± 0.04 &58.85 ± 0.05 &96.19 ± 0.00 &9.21 ± 0.08 \\
& & &MTAE &22.21 ± 0.03 &65.96 ± 0.04 &57.50 ± 0.03 &96.24 ± 0.00 &9.20 ± 0.03 \\
\cmidrule(r){2-9} 
&\multirow{3}{*}{$X^*_{\text{sig}}$} &\multirow{3}{*}{End-to-End LLaVA} &CLIP & 22.35 ± 0.03& 66.30 ± 0.05& 58.87 ± 0.06& 96.17 ± 0.00& 9.12 ± 0.06\\
& & &ViT& 21.96 ± 0.03& 65.51 ± 0.03& 58.46 ± 0.05& 96.09 ± 0.01& 9.35 ± 0.12\\
& & & SigLIP & 22.85 ± 0.05& 65.61 ± 0.01& 58.48 ± 0.03& 96.16 ± 0.00& 9.22 ± 0.11\\
\cmidrule(r){2-9} 
&\multirow{3}{*}{$X_{\text{img}}$} &\multirow{3}{*}{End-to-End LLaVA} &CLIP &21.98 ± 0.06 &66.43 ± 0.04 &59.18 ± 0.04 & 96.19 ± 0.00& 9.52 ± 0.06\\
& & &ViT&21.59 ± 0.09 & 65.13 ± 0.07&57.83 ± 0.06 & 96.05 ± 0.01& 8.83 ± 0.13\\
& & & SigLIP & 22.59 ± 0.06&65.51 ± 0.04 &58.79 ± 0.05 &96.14 ± 0.00 &9.17 ± 0.09 \\
\cmidrule(r){2-9} 
&$X_{\text{ID}}$ &End-to-End &ECG-Byte &\textbf{23.74 ± 0.05} &\textbf{67.86 ± 0.06} &\textbf{61.41 ± 0.10} &\textbf{96.30 ± 0.01} &\textbf{11.09 ± 0.10} \\

\bottomrule
\end{tabular}}}
\end{table}

\paragraph{Different LLMs}
We evaluated the effect of the LLM on each input representation and training paradigm, as summarized in Table~\ref{tab:llm}.  Specifically, we compared \mbox{Gemma-2B-Instruct}\footnote{\url{https://huggingface.co/google/gemma-2b-it}} \citep{gemmateam2024gemma2improvingopen}, \mbox{Qwen2-1.5B-Instruct}\footnote{\url{https://huggingface.co/Qwen/Qwen2-1.5B-Instruct}} \citep{qwen2025qwen25technicalreport}, and \mbox{Llama-3.2-1B-Instruct}\footnote{\url{https://huggingface.co/meta-llama/Llama-3.2-1B-Instruct}} \citep{grattafiori2024llama3herdmodels}.  
We note that for Gemma-2B-Instruct, Flash Attention 2 \citep{dao2023flashattention2fasterattentionbetter} is not utilized due to compatibility issues.
The Llama-3.2-1B results replicate those reported in Figure~\ref{fig:main_results} and Table~\ref{tab:main}.  
Table~\ref{tab:llm} shows that LLM choice has a clear impact: Llama-3.2-1B-Instruct consistently attains the highest scores across all metrics.  
Across all three LLMs, the ECG-Byte input representation yields the best performance.

\paragraph{Different Sequence Lengths}
We examine the effect of input sequence length $T$, where attention computation scales quadratically with $T$ \citep{vaswani2023attention}. We conducted an ablation study by setting $T$ to 512, 1024, and 2048, with results detailed in Table~\ref{tab:t}. For $T=1024$, these results are the same from Figure~\ref{fig:main_results} and Table~\ref{tab:main}. When $T$ is 1024 or 2048, all methods perform comparably, with ECG-Byte \citep{han2024ecgbytetokenizerendtoendgenerative} slightly outperforming the others. However, at $T=512$, ECG-Byte's performance drops significantly, whereas \textbf{2-Stage Scratch} and \textbf{End-to-End LLaVA} achieve their best performance at this sequence length. This discrepancy arises from the distinct padding schemes employed: in the End-to-End method (ECG-Byte), the number of ECG tokens $X_{\text{ID}}$ can be reduced, whereas \textbf{2-Stage Scratch} and \textbf{End-to-End LLaVA} preserve full signal information by truncating only text tokens (see Subsection~\ref{sec:pad}). Although lengths of 1024–2048 tokens are modest by modern standards, practitioners with tight computational budgets may prefer the \textbf{End-to-End LLaVA} models—they avoid training an ECG encoder from scratch and remain robust at shorter sequence lengths.

\begin{table}[!htp]\centering
\caption{Mean results over 5 random seeds with standard deviations when ablating the sequence length $T$.}\label{tab:t}
{\resizebox{\columnwidth}{!}{%
\begin{tabular}{cccccccccc}\toprule
Sequence Length $T$&Representation &Training Paradigm &$\mathcal{F}(*)$ &Bleu-4 &Rouge-L &Meteor &BertScore F1 &Accuracy \\\midrule
\multirow{11}{*}{512} &\multirow{4}{*}{$X_{\text{sig}}$} &\multirow{4}{*}{2-Stage Scratch} &MERL & 24.05 ± 0.06& 66.20 ± 0.06& 59.83 ± 0.05&96.15 ± 0.01 &9.53 ± 0.01 \\
& & &ST-MEM& 23.84 ± 0.03& 66.17 ± 0.02& 59.11 ± 0.02& 96.11 ± 0.00& 8.49 ± 0.05 \\
& & &MLAE & 23.88 ± 0.04& 66.22 ± 0.04& 59.14 ± 0.05& 96.29 ± 0.00& 9.08 ± 0.08 \\
& & &MTAE & 23.96 ± 0.03& 66.88 ± 0.04& 58.77 ± 0.03& 96.34 ± 0.00& 10.07 ± 0.03 \\
\cmidrule(r){2-9} 
&\multirow{3}{*}{$X^*_{\text{sig}}$} &\multirow{3}{*}{End-to-End LLaVA} &CLIP& 23.89 ± 0.05& \textbf{68.11 ± 0.04}& \textbf{61.49 ± 0.08}& 96.54 ± 0.01& 11.11 ± 0.12\\
& & &ViT&23.18 ± 0.06 & 68.01 ± 0.05& 60.55 ± 0.02& 96.33 ± 0.00& 11.12 ± 0.05\\
& & & SigLIP  & \textbf{24.61 ± 0.02}& 67.30 ± 0.03& 60.62 ± 0.04& 96.31 ± 0.04& \textbf{11.46 ± 0.01}\\
\cmidrule(r){2-9} 
&\multirow{3}{*}{$X_{\text{img}}$} &\multirow{3}{*}{End-to-End LLaVA} &CLIP& 23.38 ± 0.06 &67.65 ± 0.08 & \textbf{61.49 ± 0.07}& \textbf{96.68 ± 0.02}& 10.91 ± 0.11\\
& & &ViT& 23.15 ± 0.04& 67.04 ± 0.01& 58.98 ± 0.03& 96.28 ± 0.01& 10.46 ± 0.09\\
& & & SigLIP  & 24.34 ± 0.04& 66.98 ± 0.02& 60.68 ± 0.05& 96.26 ± 0.01&11.00 ± 0.10 \\
\cmidrule(r){2-9} 
&$X_{\text{id}}$ &End-to-End &ECG-Byte & 14.05 ± 0.04& 53.92 ± 0.07& 46.61 ± 0.06& 94.23 ± 0.01& 2.50 ± 0.07\\
\midrule
\multirow{11}{*}{1024} &\multirow{4}{*}{$X_{\text{sig}}$} &\multirow{4}{*}{2-Stage Scratch} &MERL &22.29 ± 0.04 &65.26 ± 0.05 & 58.52 ± 0.05& 96.08 ± 0.01&8.70 ± 0.07 \\
& & &ST-MEM& 22.09 ± 0.03& 65.26 ± 0.02 &57.83 ± 0.02 &96.01 ± 0.00 &7.75 ± 0.05 \\
& & &MLAE & 23.06 ± 0.04 &66.27 ± 0.04 &58.85 ± 0.05 &96.19 ± 0.00 &9.21 ± 0.08 \\
& & &MTAE &22.21 ± 0.03 &65.96 ± 0.04 &57.50 ± 0.03 &96.24 ± 0.00 &9.20 ± 0.03 \\
\cmidrule(r){2-9} 
&\multirow{3}{*}{$X^*_{\text{sig}}$} &\multirow{3}{*}{End-to-End LLaVA} &CLIP & 22.35 ± 0.03& 66.30 ± 0.05& 58.87 ± 0.06& 96.17 ± 0.00& 9.12 ± 0.06\\
& & &ViT& 21.96 ± 0.03& 65.51 ± 0.03& 58.46 ± 0.05& 96.09 ± 0.01& 9.35 ± 0.12\\
& & & SigLIP & 22.85 ± 0.05& 65.61 ± 0.01& 58.48 ± 0.03& 96.16 ± 0.00& 9.22 ± 0.11\\
\cmidrule(r){2-9} 
&\multirow{3}{*}{$X_{\text{img}}$} &\multirow{3}{*}{End-to-End LLaVA} &CLIP &21.98 ± 0.06 &66.43 ± 0.04 &59.18 ± 0.04 & 96.19 ± 0.00& 9.52 ± 0.06\\
& & &ViT&21.59 ± 0.09 & 65.13 ± 0.07&57.83 ± 0.06 & 96.05 ± 0.01& 8.83 ± 0.13\\
& & & SigLIP & 22.59 ± 0.06&65.51 ± 0.04 &58.79 ± 0.05 &96.14 ± 0.00 &9.17 ± 0.09 \\
\cmidrule(r){2-9} 
&$X_{\text{id}}$ &End-to-End &ECG-Byte &\textbf{23.74 ± 0.05} &\textbf{67.86 ± 0.06} &\textbf{61.41 ± 0.10} &\textbf{96.30 ± 0.01} &\textbf{11.09 ± 0.10} \\
\midrule
\multirow{11}{*}{2048} &\multirow{4}{*}{$X_{\text{sig}}$} &\multirow{4}{*}{2-Stage Scratch} &MERL & 20.30 ± 0.07& 63.62 ± 0.02& 56.35 ± 0.04& 95.88 ± 0.03& 6.75 ± 0.09\\
& & &ST-MEM& 20.12 ± 0.03& 63.63 ± 0.02& 55.69 ± 0.02& 95.81 ± 0.00& 6.01 ± 0.05 \\
& & &MLAE & 20.01 ± 0.04& 63.61 ± 0.04& 55.67 ± 0.05& 94.99 ± 0.00& 7.10 ± 0.08 \\
& & &MTAE & 20.23 ± 0.03& 64.31 ± 0.04& 55.37 ± 0.03& 96.04 ± 0.00& 7.14 ± 0.03 \\
\cmidrule(r){2-9} 
&\multirow{3}{*}{$X^*_{\text{sig}}$} &\multirow{3}{*}{End-to-End LLaVA} &CLIP& 21.92 ± 0.04&65.35 ± 0.05 &57.57 ± 0.05 & 95.67 ± 0.01& 7.61 ± 0.05 \\
& & &ViT& 21.05 ± 0.03& 63.82 ± 0.04& 56.09 ± 0.05& 95.77 ± 0.01& 7.10 ± 0.03\\
& & & SigLIP  & 22.01 ± 0.02& 64.56 ± 0.05&57.83 ± 0.04 & 96.03 ± 0.02&7.69 ± 0.04 \\
\cmidrule(r){2-9} 
&\multirow{3}{*}{$X_{\text{img}}$} &\multirow{3}{*}{End-to-End LLaVA} &CLIP& 21.96 ± 0.06&65.43 ± 0.03 &57.68 ± 0.05& 96.21 ± 0.02 & 7.89 ± 0.06\\
& & &ViT&21.29 ± 0.05 & 64.31 ± 0.04& 56.50 ± 0.06& 95.94 ± 0.01&7.31 ± 0.03 \\
& & & SigLIP  & 22.10 ± 0.04& 64.72 ± 0.02&57.93 ± 0.01 & 96.04 ± 0.03& 7.66 ± 0.05\\
\cmidrule(r){2-9} 
&$X_{\text{id}}$ &End-to-End &ECG-Byte & \textbf{23.56 ± 0.03}& \textbf{67.78 ± 0.04}& \textbf{60.00 ± 0.03}& \textbf{96.33 ± 0.00}& \textbf{11.01 ± 0.02}\\
\bottomrule
\end{tabular}}}
\end{table}

\paragraph{Different ECG Lengths}
The ECG length \(L\) denotes the number of sampled timesteps kept from each ECG recording. After resampling all traces to 250 Hz, \(L = 500\), \(1250\), and \(2500\) correspond to 2, 5, and the full 10 second segments, respectively. Clinical experts require at least 2 seconds of signal for a reliable reading; longer windows add more contextual information. We therefore benchmark at these three lengths, re-using the \(L = 1250\) setting from Figure~\ref{fig:main_results} and Table~\ref{tab:main}. As shown in Table~\ref{tab:l}, ECG-Byte consistently outperforms all baselines, with its advantage growing as \(L\) increases. This is intuitive since \textbf{2-Stage Scratch} and \textbf{End-to-End LLaVA} methods compress the ECG into one token, while ECG-Byte compresses the signal while maintaining the length.
\newpage
\begin{table}[!htp]\centering
\caption{Mean results over 5 random seeds with standard deviations when ablating the ECG length $L$.}\label{tab:l}
{\resizebox{0.9\columnwidth}{!}{%
\begin{tabular}{cccccccccc}\toprule
ECG Length $L$ &Representation &Training Paradigm &$\mathcal{F}(*)$ &Bleu-4 &Rouge-L &Meteor &BertScore F1 &Accuracy \\\midrule
\multirow{11}{*}{500} &\multirow{4}{*}{$X_{\text{sig}}$} &\multirow{4}{*}{2-Stage Scratch} &MERL & 22.17 ± 0.07& 66.32 ± 0.06 &58.17 ± 0.04 &96.06 ± 0.01 &8.84 ± 0.05 \\
& & & ST-MEM & 21.97 ± 0.03 & 66.32 ± 0.02 & 57.47 ± 0.02 & 96.00 ± 0.00 & 7.88 ± 0.05 \\
& & & MLAE  & 22.95 ± 0.04 & 67.31 ± 0.04 & 58.52 ± 0.05 & 96.12 ± 0.00 & 9.36 ± 0.08 \\
& & & MTAE  & 22.08 ± 0.03 & 67.05 ± 0.04 & 57.19 ± 0.03 & \textbf{96.24 ± 0.00} & 9.34 ± 0.03 \\
\cmidrule(r){2-9} 
&\multirow{3}{*}{$X^*_{\text{sig}}$} &\multirow{3}{*}{End-to-End LLaVA} &CLIP  &22.41 ± 0.06& 65.91 ± 0.03&59.16 ± 0.05 &96.10 ± 0.02 &8.70 ± 0.04 \\
& & &ViT & 23.19 ± 0.04& 66.41 ± 0.04&59.22 ± 0.06 &96.08 ± 0.01 &9.66 ± 0.05 \\
& & & SigLIP & 23.14 ± 0.03& 66.29 ± 0.05& 59.10 ± 0.04& 96.10 ± 0.01& 9.32 ± 0.04\\
\cmidrule(r){2-9} 
&\multirow{3}{*}{$X_{\text{img}}$} &\multirow{3}{*}{End-to-End LLaVA} &CLIP  & 22.03 ± 0.06 & 66.04 ± 0.04 & 59.46 ± 0.04 & 96.10 ± 0.00 & 9.08 ± 0.06 \\
& & & ViT & 22.80 ± 0.09 & 66.03 ± 0.07 & 58.57 ± 0.06 & 96.00 ± 0.01 & 9.12 ± 0.13 \\
& & & SigLIP & 22.88 ± 0.06 & 66.17 ± 0.04 & 59.40 ± 0.05 & 96.00 ± 0.00 & 9.27 ± 0.09 \\
\cmidrule(r){2-9} 
&$X_{\text{id}}$ &End-to-End &ECG-Byte &\textbf{23.45 ± 0.05} & \textbf{67.53 ± 0.02}&\textbf{62.38 ± 0.04} & 96.23 ± 0.01&\textbf{11.24 ± 0.03} \\
\midrule
\multirow{11}{*}{1250} &\multirow{4}{*}{$X_{\text{sig}}$} &\multirow{4}{*}{2-Stage Scratch} &MERL &22.29 ± 0.04 &65.26 ± 0.05 & 58.52 ± 0.05& 96.08 ± 0.01&8.70 ± 0.07 \\
& & &ST-MEM& 22.09 ± 0.03& 65.26 ± 0.02 &57.83 ± 0.02 &96.01 ± 0.00 &7.75 ± 0.05 \\
& & &MLAE & 23.06 ± 0.04 &66.27 ± 0.04 &58.85 ± 0.05 &96.19 ± 0.00 &9.21 ± 0.08 \\
& & &MTAE &22.21 ± 0.03 &65.96 ± 0.04 &57.50 ± 0.03 &96.24 ± 0.00 &9.20 ± 0.03 \\
\cmidrule(r){2-9} 
&\multirow{3}{*}{$X^*_{\text{sig}}$} &\multirow{3}{*}{End-to-End LLaVA} &CLIP & 22.35 ± 0.03& 66.30 ± 0.05& 58.87 ± 0.06& 96.17 ± 0.00& 9.12 ± 0.06\\
& & &ViT& 21.96 ± 0.03& 65.51 ± 0.03& 58.46 ± 0.05& 96.09 ± 0.01& 9.35 ± 0.12\\
& & & SigLIP & 22.85 ± 0.05& 65.61 ± 0.01& 58.48 ± 0.03& 96.16 ± 0.00& 9.22 ± 0.11\\
\cmidrule(r){2-9} 
&\multirow{3}{*}{$X_{\text{img}}$} &\multirow{3}{*}{End-to-End LLaVA} &CLIP &21.98 ± 0.06 &66.43 ± 0.04 &59.18 ± 0.04 & 96.19 ± 0.00& 9.52 ± 0.06\\
& & &ViT&21.59 ± 0.09 & 65.13 ± 0.07&57.83 ± 0.06 & 96.05 ± 0.01& 8.83 ± 0.13\\
& & & SigLIP & 22.59 ± 0.06&65.51 ± 0.04 &58.79 ± 0.05 &96.14 ± 0.00 &9.17 ± 0.09 \\
\cmidrule(r){2-9} 
&$X_{\text{id}}$ &End-to-End &ECG-Byte &\textbf{23.74 ± 0.05} &\textbf{67.86 ± 0.06} &\textbf{61.41 ± 0.10} &\textbf{96.30 ± 0.01} &\textbf{11.09 ± 0.10} \\
\midrule
\multirow{11}{*}{2500} &\multirow{4}{*}{$X_{\text{sig}}$} &\multirow{4}{*}{2-Stage Scratch} &MERL & 22.78 ± 0.04 & 65.59 ± 0.05 & 58.52 ± 0.05 & 96.08 ± 0.01 & 8.06 ± 0.07 \\
& & & ST-MEM & 22.58 ± 0.03 & 65.59 ± 0.02 & 57.83 ± 0.02 & 96.01 ± 0.00 & 7.18 ± 0.05 \\
& & & MLAE  & 22.57 ± 0.04 & 65.60 ± 0.04 & 57.85 ± 0.05 & 96.19 ± 0.00 & 8.54 ± 0.08 \\
& & & MTAE  & 22.70 ± 0.03 & 66.29 ± 0.04 & 57.50 ± 0.03 & 96.24 ± 0.00 & 8.53 ± 0.03 \\
\cmidrule(r){2-9} 
&\multirow{3}{*}{$X^*_{\text{sig}}$} &\multirow{3}{*}{End-to-End LLaVA} &CLIP  & 22.55 ± 0.10 & 66.00 ± 0.05&58.85 ± 0.13 & 95.31 ± 0.02& 8.57 ± 0.11\\
& & &ViT & 22.51 ± 0.05 & 66.63 ± 0.08& 58.64 ± 0.13&96.27 ± 0.01 &8.38 ± 0.05 \\
& & & SigLIP & 22.67 ± 0.01& 65.68 ± 0.06 &58.41 ± 0.03 &96.21 ± 0.02 &8.47 ± 0.13 \\
\cmidrule(r){2-9} 
&\multirow{3}{*}{$X_{\text{img}}$} &\multirow{3}{*}{End-to-End LLaVA} &CLIP  & 22.84 ± 0.12 & 66.09 ± 0.08& 58.96 ± 0.09 & 96.18 ± 0.02& 8.51 ± 0.08\\
& & &ViT & 22.49 ± 0.01& 66.79 ± 0.02&58.81 ± 0.00 & 96.28 ± 0.00& 8.45 ± 0.01\\
& & & SigLIP &22.43 ± 0.02 & 65.42 ± 0.03&57.84 ± 0.06 & 96.19 ± 0.02&8.34 ± 0.06 \\
\cmidrule(r){2-9} 
&$X_{\text{id}}$ &End-to-End &ECG-Byte & \textbf{25.09 ± 0.09} & \textbf{68.06 ± 0.07}& \textbf{61.63 ± 0.10}& \textbf{96.40 ± 0.01} & \textbf{10.78 ± 0.13}\\
\bottomrule
\end{tabular}}}
\end{table}

\section{Discussion and Conclusion}
Our comprehensive benchmark reveals clear, statistically significant advantages of the tokenized ECG representation $X_{\text{ID}}$ when trained End-to-End, compared to both raw signal ($X_{\text{sig}}$, $X^*_{\text{sig}}$) and image ($X_{\text{img}}$) modalities. Across six datasets and five evaluation metrics, ECG-Byte’s \textbf{End-to-End} approach consistently achieves the highest mean scores and secures the greatest number of significant wins in two-sample $z$-tests (Tables~\ref{tab:overall_wins}). This performance gap widens with longer ECG segments ($L=2500$) and remains robust under realistic perturbations (Table~\ref{tab:ecg_chat_instruct_Perturbed}), suggesting that the model effectively leverages compressed symbolic information even when signal fidelity degrades.

Our ablation studies further illuminate key trade-offs. While End-to-End tokenization excels at standard and extended sequence lengths ($T\ge1024$), it suffers at $T=512$ due to aggressive ECG token truncation, whereas \textbf{End-to-End LLaVA} models maintain stable performance by preserving full signal context (Table~\ref{tab:t}). This indicates that practitioners constrained by memory or latency may favor pretrained encoder pipelines, trading some generative quality for sequence-length resilience.

In conclusion, our unified framework demonstrates that symbolic, tokenized representations of ECGs constitute the most effective input modality for autoregressive ELMs, achieving superior generative performance, strong robustness, and efficient scaling. We release our codebase and pretrained components to facilitate reproducible research and encourage the community to build on these findings. By standardizing input representations, future ELM development can more confidently allocate resources toward model scaling and clinical translation, ultimately advancing automated, interpretable ECG analysis.

\impact{Our work on providing a comprehensive benchmark for different input representations of ECGs for further processing into Electrocardiogram-Language Models (ELMs) is purposed with the goal to create the world's first expert, artifical cardiac electrophysiologist. To get to this goal, we provide an open-source, comprehensive evaluation of different input representations of ECGs for ELMs. This work, however, has some potential negative societal consequences as well. Although we present a comprehensive evaluation framework for modern ELMS, we caution researchers and even clinical practitioners to note that the performances of even the top performing input representation is nowhere near the skill of an expert cardiac electrophysiologist. Much further development in model architecture as well as large scale training is needed for further improvement. We hope that our open-source, unified framework and comprehensive evaluations provides researchers a guideline in developing the next generation ELMs.}


\acks{This work is done in collaboration with the Mario
Lemieux Center for Heart Rhythm Care at Allegheny
General Hospital.}

\vskip 0.2in
\bibliography{sample}
\newpage
\appendix
\section{System Prompt for ELMs}
\label{sys_prompt}
\begin{tcolorbox}[colback=white, colframe=black, sharp corners, boxrule=0.5mm]
You are an expert multimodal assistant capable of processing both natural language text and ECG signals.
When you receive input, first determine if it is text, ECG data, or both. For ECG signals, interpret them as
time-series data representing cardiac activity—analyzing features such as heart rate, rhythm, and potential abnormalities.
When both modalities are present, synthesize the information to provide integrated, expert cardiac electrophysiologist-level responses.
Your answers should be precise, concise, and informed by clinical signal analysis and natural language understanding.
Additionally, if the user asks a general question, you should answer it as a general assistant.
\end{tcolorbox}

\section{Input Templates}
\label{input_temp}

We provide the input template for each LLM used in our study. We utilize the conversational templates provided by FastChat \citep{zheng2023judging}.

\begin{tcolorbox}[colback=white, colframe=black, sharp corners, boxrule=0.5mm]
\textbf{Llama 3.2 1B Instruct Conversation Template}\\\\
$<|\text{begin\_of\_text}|><|\text{start\_header\_id}|>\text{system}<|\text{end\_header\_id}|>$\\\\
$q_{\text{sys}}<|\text{eot\_id}|><|\text{start\_header\_id}|>\text{user}<|\text{end\_header\_id}|>$\\\\
$<\text{signal}>$\\
$q_1<|\text{eot\_id}|><|\text{start\_header\_id}|>\text{assistant}<|\text{end\_header\_id}|>$\\\\

$s_1<|\text{eot\_id}|><|\text{start\_header\_id}|>\text{user}<|\text{end\_header\_id}|>$\\\\

$q_2<|\text{eot\_id}|><|\text{start\_header\_id}|>\text{assistant}<|\text{end\_header\_id}|>$\\\\

$s_2<|\text{eot\_id}|>$

... \\

$q_n<|\text{eot\_id}|><|\text{start\_header\_id}|>\text{assistant}<|\text{end\_header\_id}|>$\\\\

$s_n<|\text{eot\_id}|>$
\end{tcolorbox}

\begin{tcolorbox}[colback=white, colframe=black, sharp corners, boxrule=0.5mm]
\textbf{Gemma 2 2B Instruct Conversation Template}\\\\
$<\text{bos}><\text{start\_of\_turn}>\text{user}$\\
$<\text{signal}>$\\
$q_1<\text{end\_of\_turn}>$\\
$<\text{start\_of\_turn}>\text{model}$\\
$s_1<\text{end\_of\_turn}>$\\
$<\text{start\_of\_turn}>\text{user}$\\
$q_2<\text{end\_of\_turn}>$\\
$<\text{start\_of\_turn}>\text{model}$\\
$s_2<\text{end\_of\_turn}>$\\

... \\

$<\text{start\_of\_turn}>\text{user}$\\
$q_n<\text{end\_of\_turn}>$\\
$<\text{start\_of\_turn}>\text{model}$\\
$s_n<\text{end\_of\_turn}>$\\
\end{tcolorbox}

\begin{tcolorbox}[colback=white, colframe=black, sharp corners, boxrule=0.5mm]
\textbf{Qwen 2.5 1.5B Instruct Conversation Template}\\\\
$<|\text{im\_start}|>\text{system}$\\
$q_{\text{sys}}<|\text{im\_end}|>$\\
$<|\text{im\_start}|>\text{user}$\\
$<\text{signal}>$\\
$q_1<|\text{im\_end}|>$\\
$<|\text{im\_start}|>\text{assistant}$\\
$s_1<|\text{im\_end}|>$\\
$<|\text{im\_start}|>\text{user}$\\
$q_2<|\text{im\_end}|>$\\
$<|\text{im\_start}|>\text{assistant}$\\
$s_2<|\text{im\_end}|>$\\

... \\

$<|\text{im\_start}|>\text{user}$\\
$q_n<|\text{im\_end}|>$\\
$<|\text{im\_start}|>\text{assistant}$\\
$s_n<|\text{im\_end}|>$\\
\end{tcolorbox}

\section{Additional Results}
\label{apd:results}
\begin{table*}[!ht]
\centering
\caption{Baseline vs. perturbation averaged performances over 5 random seeds on the ECG-Chat Instruct dataset. $\Delta$ is calculated as Perturbed – Baseline.}
\label{tab:ecg_chat_instruct_Perturbed_main}
\resizebox{\textwidth}{!}{%
\begin{tabular}{lll
                *{3}{c} *{3}{c} *{3}{c} *{3}{c} *{3}{c}}
\toprule
\multirow{2}{*}{Representation}
  & \multirow{2}{*}{Training Paradigm}
  & \multirow{2}{*}{$\mathcal{F}(*)$}
  & \multicolumn{3}{c}{BLEU-4}
  & \multicolumn{3}{c}{ROUGE-L}
  & \multicolumn{3}{c}{METEOR}
  & \multicolumn{3}{c}{BERTScore F1}
  & \multicolumn{3}{c}{Accuracy} \\
\cmidrule(lr){4-6}\cmidrule(lr){7-9}\cmidrule(lr){10-12}
\cmidrule(lr){13-15}\cmidrule(lr){16-18}
 & & & Baseline & Perturbed & $\Delta$ & Baseline & Perturbed & $\Delta$
       & Baseline & Perturbed & $\Delta$ & Baseline & Perturbed & $\Delta$
       & Baseline & Perturbed & $\Delta$ \\
\midrule
\multirow{4}{*}{$X_{\text{sig}}$}
 & \multirow{4}{*}{2-Stage Scratch}
 & MERL  & 22.29 ± 0.04 & 22.22 ± 0.05 & -0.07 & 65.26 ± 0.05 & 65.21 ± 0.06 & -0.05 &
           58.52 ± 0.05 & 58.44 ± 0.04 & -0.08 & 96.08 ± 0.01 & 96.00 ± 0.00 & -0.08 &
           8.70 ± 0.07  & 8.75 ± 0.09 & +0.05 \\[2pt]
 & & ST-MEM & 22.09 ± 0.03 & 22.13 ± 0.05 & +0.04 & 65.26 ± 0.02 & 65.23 ± 0.01 & -0.03 &
               57.83 ± 0.02 & 57.82 ± 0.05 & -0.01 & 96.01 ± 0.00 & 96.00 ± 0.02 & -0.01 &
               7.75 ± 0.05 & 7.76 ± 0.04 & +0.01 \\[2pt]
 & & MLAE  & 23.06 ± 0.04 & 23.07 ± 0.05 & +0.01 & 66.27 ± 0.04 & 66.20 ± 0.06 & -0.07 &
              58.85 ± 0.05 & 58.80 ± 0.04 & -0.05 & 96.19 ± 0.00 & 96.10 ± 0.02 & -0.09 &
              9.21 ± 0.08 & 9.12 ± 0.09 & -0.09 \\[2pt]
 & & MTAE  & 22.21 ± 0.03 & 22.17 ± 0.04 & -0.04 & 65.96 ± 0.04 & 65.99 ± 0.07 & \textbf{+0.03} &
              57.50 ± 0.03 & 57.54 ± 0.05 & \textbf{+0.04} & 96.24 ± 0.00 & 96.20 ± 0.01 & -0.04 &
              9.20 ± 0.03 & 9.30 ± 0.05 & +0.10 \\
\cmidrule(r){1-18}
\multirow{3}{*}{$X^{*}_{\text{sig}}$}
 & \multirow{3}{*}{End-to-End LLaVA}
 & CLIP  & 22.35 ± 0.03 & 22.38 ± 0.04 & +0.03 & 66.30 ± 0.05 & 66.30 ± 0.06 & 0.00 &
            58.87 ± 0.06 & 58.84 ± 0.05 & -0.03 & 96.17 ± 0.00 & 96.17 ± 0.01 & \textbf{0.00} &
            9.12 ± 0.06 & 9.23 ± 0.08 & +0.11 \\[2pt]
 & & ViT   & 21.96 ± 0.03 & 22.01 ± 0.04 & +0.05 & 65.51 ± 0.03 & 65.51 ± 0.04 & 0.00 &
             58.46 ± 0.05 & 58.49 ± 0.03 & +0.03 & 96.09 ± 0.01 & 96.09 ± 0.02 & \textbf{0.00 }&
             9.35 ± 0.12 & 9.43 ± 0.15 & +0.08 \\[2pt]
 & & SigLIP & 22.85 ± 0.05 & 22.92 ± 0.05 & +0.07 & 65.61 ± 0.01 & 65.57 ± 0.06 & -0.04  &
               58.48 ± 0.03 & 58.40 ± 0.07 & -0.08  & 96.16 ± 0.00 & 96.15 ± 0.01 & -0.01  &
               9.22 ± 0.11 & 9.35 ± 0.12 & \textbf{+0.13} \\
\cmidrule(r){1-18}
\multirow{3}{*}{$X_{\text{img}}$}
 & \multirow{3}{*}{End-to-End LLaVA}
 & CLIP  & 21.98 ± 0.06 & 21.88 ± 0.05 & -0.10 & 66.43 ± 0.04 & 66.41 ± 0.05 & -0.02 &
            59.18 ± 0.04 & 59.12 ± 0.05 & -0.06 & 96.19 ± 0.00 & 96.19 ± 0.01 & \textbf{0.00} &
            9.52 ± 0.06 & 9.40 ± 0.03 & -0.12 \\[2pt]
 & & ViT   & 21.59 ± 0.09 & 21.58 ± 0.08 & -0.01 & 65.13 ± 0.07 & 65.10 ± 0.06 & -0.03 &
             57.83 ± 0.06 & 57.75 ± 0.05 & -0.08 & 96.05 ± 0.01 & 96.04 ± 0.00 & -0.01 &
             8.83 ± 0.13 & 8.77 ± 0.14 & -0.06 \\[2pt]
 & & SigLIP & 22.59 ± 0.06 & 22.58 ± 0.07 & -0.01 & 65.51 ± 0.04 & 65.48 ± 0.03 & -0.03 &
               58.79 ± 0.05 & 58.81 ± 0.06 & +0.02 & 96.14 ± 0.00 & 96.14 ± 0.01 & \textbf{0.00} &
               9.17 ± 0.09 & 9.13 ± 0.06 & -0.04 \\
\cmidrule(r){1-18}
$X_{\text{id}}$ & End-to-End & ECG-Byte
 & 23.74 ± 0.05 & 23.83 ± 0.07 & \textbf{+0.09} & 67.86 ± 0.06 & 67.83 ± 0.05 & -0.03 &
   61.41 ± 0.10 & 61.43 ± 0.07 & +0.02 & 96.30 ± 0.01 & 96.30 ± 0.01 & \textbf{0.00} &
   11.09 ± 0.10 & 11.04 ± 0.11 & -0.05 \\
\bottomrule
\end{tabular}}
\end{table*}

\begin{table}[!htp]\centering
\caption{Numerical mean results over five random seeds with standard deviations of the spider chart in Figure~\ref{fig:main_results}.}\label{tab:main}
{\resizebox{\columnwidth}{!}{%
\begin{tabular}{lccccccccc}\toprule
Training/Inference Dataset &Representation &Training Paradigm &$\mathcal{F}(*)$ &Bleu-4 &Rouge-L &Meteor &BertScore F1 &Accuracy \\\midrule
\multirow{11}{*}{ECG-Chat Pretrain} &\multirow{4}{*}{$X_{\text{sig}}$} &\multirow{4}{*}{2-Stage Scratch} &MERL & 4.77 ± 0.06 &\textbf{44.33 ± 0.11} &\textbf{26.86 ± 0.16} &\textbf{93.06 ± 0.02} &\textbf{11.12 ± 0.10} \\
& & &ST-MEM& 6.93 ± 0.04 & 32.83 ± 0.05 & 23.83 ± 0.08 &90.82 ± 0.01 &0.58 ± 0.09 \\
& & &MLAE &7.64 ± 0.10 &36.28 ± 0.07 &24.35 ± 0.08 &91.65 ± 0.01 & 3.67 ± 0.11 \\
& & &MTAE &5.71 ± 0.08 &31.36 ± 0.14 &20.09 ± 0.18 &90.90 ± 0.02 &1.61 ± 0.12 \\
\cmidrule(r){2-9} 
&\multirow{3}{*}{$X^*_{\text{sig}}$} &\multirow{3}{*}{End-to-End LLaVA} &CLIP & 3.56 ± 0.03 &37.99 ± 0.14 &19.89 ± 0.15 &92.03 ± 0.01 &6.23 ± 0.18 \\
& & &ViT& 3.33 ± 0.06 & 38.47 ± 0.11 &20.58 ± 0.13 &92.15 ± 0.01 &6.09 ± 0.14 \\
& & & SigLIP & 6.29 ± 0.07 & 34.31 ± 0.05 &23.08 ± 0.08 &91.35 ± 0.01 &2.24 ± 0.07 \\
\cmidrule(r){2-9} 
&\multirow{3}{*}{$X_{\text{img}}$} &\multirow{3}{*}{End-to-End LLaVA} &CLIP & 3.85 ± 0.04 & 40.16 ± 0.08 &23.27 ± 0.11 & 92.51 ± 0.01 & 7.67 ± 0.08 \\
& & &ViT&4.42 ± 0.04 & 38.85 ± 0.15 & 23.10 ± 0.18 &92.34 ± 0.02 &6.72 ± 0.16 \\
& & & SigLIP &7.42 ± 0.08 &32.23 ± 0.12 &22.96 ± 0.11 &90.87 ± 0.02 &0.39 ± 0.04 \\
\cmidrule(r){2-9} 
&$X_{\text{id}}$ &End-to-End &ECG-Byte & \textbf{7.72 ± 0.12} & 36.46 ± 0.11 & 24.66 ± 0.15 & 91.50 ± 0.02 & 3.98 ± 0.14 \\
\midrule
\multirow{11}{*}{ECG-Chat Instruct} &\multirow{4}{*}{$X_{\text{sig}}$} &\multirow{4}{*}{2-Stage Scratch} &MERL &22.29 ± 0.04 &65.26 ± 0.05 &58.52 ± 0.05 &96.08 ± 0.01 &8.70 ± 0.07 \\
& & &ST-MEM& 22.09 ± 0.03 &65.26 ± 0.02 &57.83 ± 0.02 &96.01 ± 0.00 &7.75 ± 0.05 \\
& & &MLAE &23.06 ± 0.04 &66.27 ± 0.04 &58.85 ± 0.05 &96.19 ± 0.00 &9.21 ± 0.08 \\
& & &MTAE &22.21 ± 0.03 &65.96 ± 0.04 &57.50 ± 0.03 &96.24 ± 0.00 &9.20 ± 0.03 \\
\cmidrule(r){2-9} 
&\multirow{3}{*}{$X^*_{\text{sig}}$} &\multirow{3}{*}{End-to-End LLaVA} &CLIP &22.35 ± 0.03 &66.30 ± 0.05 &58.87 ± 0.06 &96.17 ± 0.00 &9.12 ± 0.06 \\
& & &ViT&21.96 ± 0.03 &65.51 ± 0.03 &58.46 ± 0.05 &96.09 ± 0.01 &9.35 ± 0.12 \\
& & &SigLIP &22.85 ± 0.05 &65.61 ± 0.01 &58.48 ± 0.03 &96.16 ± 0.00 &9.22 ± 0.11 \\
\cmidrule(r){2-9} 
&\multirow{3}{*}{$X_{\text{img}}$} &\multirow{3}{*}{End-to-End LLaVA} &CLIP &21.98 ± 0.06 &66.43 ± 0.04 &59.18 ± 0.04 &96.19 ± 0.00 &9.52 ± 0.06 \\
& & &ViT&21.59 ± 0.09 &65.13 ± 0.07 &57.83 ± 0.06 &96.05 ± 0.01 &8.83 ± 0.13 \\
& & &SigLIP &22.59 ± 0.06 &65.51 ± 0.04 &58.79 ± 0.05 &96.14 ± 0.00 &9.17 ± 0.09 \\
\cmidrule(r){2-9} 
&$X_{\text{id}}$ &End-to-End &ECG-Byte &\textbf{23.74 ± 0.05} &\textbf{67.86 ± 0.06} &\textbf{61.41 ± 0.10} &\textbf{96.30 ± 0.01} &\textbf{11.09 ± 0.10} \\
\midrule
\multirow{11}{*}{ECG-QA PTB-XL} &\multirow{4}{*}{$X_{\text{sig}}$} &\multirow{4}{*}{2-Stage Scratch} &MERL & 6.77 ± 0.09 &40.12 ± 0.19 &23.00 ± 0.08 &96.47 ± 0.01 &36.51 ± 0.18 \\
& & &ST-MEM& 2.77 ± 0.15 &42.97 ± 0.13 &23.37 ± 0.10 &96.54 ± 0.01 &\textbf{39.18 ± 0.14} \\
& & &MLAE & 5.11 ± 0.07 &41.59 ± 0.08 &23.70 ± 0.03 &96.59 ± 0.01 &36.68 ± 0.04 \\
& & &MTAE &13.03 ± 0.18 &43.80 ± 0.12 &27.09 ± 0.13 &96.41 ± 0.01 &36.79 ± 0.08 \\
\cmidrule(r){2-9} 
&\multirow{3}{*}{$X^*_{\text{sig}}$} &\multirow{3}{*}{End-to-End LLaVA} &CLIP &4.15 ± 0.13 &40.31 ± 0.29 &22.29 ± 0.19 &96.43 ± 0.01 &37.48 ± 0.29 \\
& & &ViT&6.02 ± 0.13 &42.32 ± 0.24 &24.17 ± 0.14 &96.54 ± 0.02 &37.09 ± 0.18 \\
& & &SigLIP &3.64 ± 0.07 &42.20 ± 0.17 &23.34 ± 0.09 &96.48 ± 0.01 &38.31 ± 0.22 \\
\cmidrule(r){2-9} 
&\multirow{3}{*}{$X_{\text{img}}$} &\multirow{3}{*}{End-to-End LLaVA} &CLIP &2.31 ± 0.10 &43.26 ± 0.13 &23.56 ± 0.12 &96.56 ± 0.01 &39.86 ± 0.10 \\
& & &ViT&6.00 ± 0.06 &42.15 ± 0.17 &24.28 ± 0.10 &\textbf{96.61 ± 0.01} &37.52 ± 0.11 \\
& & &SigLIP &3.04 ± 0.11 &37.78 ± 0.10 &20.90 ± 0.05 &96.50 ± 0.01 &33.62 ± 0.09 \\
\cmidrule(r){2-9} 
&$X_{\text{id}}$ &End-to-End &ECG-Byte &\textbf{17.83 ± 0.28} &\textbf{44.85 ± 0.28} &\textbf{28.95 ± 0.23} &96.57 ± 0.01 &36.25 ± 0.28 \\
\midrule
\multirow{11}{*}{ECG-QA MIMIC-IV-ECG} &\multirow{4}{*}{$X_{\text{sig}}$} &\multirow{4}{*}{2-Stage Scratch} &MERL &19.77 ± 0.15 &37.09 ± 0.24 &\textbf{25.69 ± 0.19 }&95.06 ± 0.02 &24.66 ± 0.26 \\
& & &ST-MEM&16.69 ± 0.12 &33.67 ± 0.21 &22.97 ± 0.14 &95.15 ± 0.02 &21.68 ± 0.21 \\
& & &MLAE &19.70 ± 0.12 &34.62 ± 0.15 &24.43 ± 0.09 &95.08 ± 0.01 &21.07 ± 0.14 \\
& & &MTAE &15.00 ± 0.19 &37.44 ± 0.17 &24.23 ± 0.12 &95.01 ± 0.02 &26.55 ± 0.12 \\
\cmidrule(r){2-9} 
&\multirow{3}{*}{$X^*_{\text{sig}}$} &\multirow{3}{*}{End-to-End LLaVA} &CLIP &12.84 ± 0.09 &35.02 ± 0.32 &22.89 ± 0.21 &94.56 ± 0.02 &24.42 ± 0.23 \\
& & &ViT&16.44 ± 0.18 &35.61 ± 0.41 &23.95 ± 0.29 &94.84 ± 0.02 &23.87 ± 0.37 \\
& & &SigLIP &13.02 ± 0.13 &33.26 ± 0.22 &22.68 ± 0.10 &94.71 ± 0.02 &22.45 ± 0.21 \\
\cmidrule(r){2-9} 
&\multirow{3}{*}{$X_{\text{img}}$} &\multirow{3}{*}{End-to-End LLaVA} &CLIP &\textbf{19.92 ± 0.15} &37.82 ± 0.21 &25.30 ± 0.09 &95.15 ± 0.01 &25.38 ± 0.20 \\
& & &ViT&9.99 ± 0.10 &36.50 ± 0.09 &22.42 ± 0.05 &95.12 ± 0.01 &\textbf{27.20 ± 0.16} \\
& & &SigLIP &14.55 ± 0.12 &29.86 ± 0.21 &20.21 ± 0.05 &\textbf{95.50 ± 0.01} &19.27 ± 0.15 \\
\cmidrule(r){2-9} 
&$X_{\text{id}}$ &End-to-End &ECG-Byte &16.47 ± 0.44 &\textbf{39.05 ± 0.13} &24.94 ± 0.16 &95.35 ± 0.02 &27.18 ± 0.13 \\
\bottomrule
\end{tabular}}}
\end{table}

\end{document}